\documentclass[preprint,3p,twocolumn]{elsarticle}

\usepackage{amsmath,amssymb,amsfonts}
\usepackage{graphicx}
\usepackage{algorithm}
\usepackage{algpseudocode}
\usepackage{enumitem}
\usepackage{balance} 
\usepackage{textcomp}
\usepackage{xcolor}
\usepackage{lipsum}
\usepackage{booktabs}
\usepackage{subcaption}
\usepackage{mwe}
\usepackage{adjustbox}
\usepackage{breqn}
\usepackage{fancyhdr}
\usepackage{multirow}
\usepackage{caption}
\usepackage{csquotes}
\usepackage{colortbl}
\usepackage{comment}

\begin{document}

\begin{frontmatter}

\title{Backpropagation-free Spiking Neural Networks with the Forward-Forward Algorithm} %% Article title

\author[cssbu]{Mohammadnavid Ghader} %% Author name
\ead{m\_ghader@sbu.ac.ir}
\author[cssbu]{Saeed Reza Kheradpisheh\corref{cor1}} %% Author name
\ead{s\_kheradpisheh@sbu.ac.ir}

\author[cybsbu]{Bahar Farahani} %% Author name
\ead{b\_farahani@sbu.ac.ir}
\author[Hertfordshire]{Mahmood Fazlali} %% Author name
\ead{m.fazlali@herts.ac.uk}

\affiliation[cssbu]{Department of Computer and Data Science, Faculty of Mathematical Sciences, Shahid Beheshti University, Tehran, Iran }
\affiliation[cybsbu]{Cyberspace Research Institute, Shahid Beheshti University, Tehran, Iran }
\affiliation[Hertfordshire]{Cybersecurity and Computing Systems Research Group, University of Hertfordshire, Hatfield, United Kingdom}
\cortext[cor1]{Corresponding authors}

%% Abstract
\begin{abstract}
Spiking Neural Networks (SNNs) offer a biologically inspired computational paradigm that emulates neuronal activity through discrete spike-based processing. Despite their advantages, training SNNs with traditional backpropagation (BP) remains challenging due to computational inefficiencies and a lack of biological plausibility. This study explores the Forward-Forward (FF) algorithm as an alternative learning framework for SNNs. Unlike backpropagation, which relies on forward and backward passes, the FF algorithm employs two forward passes, enabling layer-wise localized learning, enhanced computational efficiency, and improved compatibility with neuromorphic hardware. We introduce an FF-based SNN training framework and evaluate its performance across both non-spiking (MNIST, Fashion-MNIST, Kuzushiji-MNIST) and spiking (Neuro-MNIST, SHD) datasets. Experimental results demonstrate that our model surpasses existing FF-based SNNs on evaluated static datasets with a much lighter architecture while achieving accuracy comparable to state-of-the-art backpropagation-trained SNNs. On more complex spiking tasks such as SHD, our approach outperforms other SNN models and remains competitive with leading backpropagation-trained SNNs. These findings highlight the FF algorithm’s potential to advance SNN training methodologies by addressing some key limitations of backpropagation.
\end{abstract}

%% Keywords
\begin{keyword}
Spiking Neural Network\sep Forward-Forward Algorithm\sep Forward-Only Learning\sep Backpropagation-free

\end{keyword}

\end{frontmatter}

\section{Introduction}
\label{sec1}
Spiking Neural Networks (SNNs), as a biologically inspired model of neural computation, are designed to process information using discrete spike events~\cite{b17}, closely mimicking the dynamics of real neurons. However, training SNNs remains a significant challenge, primarily due to the credit assignment problem~\cite{b20} and the difficulty of attributing the contributions of individual neurons to network output errors. Unlike traditional artificial neural networks, SNNs operate on non-differentiable spiking activities, complicating the application of gradient-based training methods like backpropagation~\cite{snnbp}. To address this, surrogate gradient~\cite{neftci2019} methods are introduced, allowing approximate gradients for effective training while preserving the spike-based behavior of SNNs. The surrogate gradient method replaces the non-differentiable spike activation function with a smooth, differentiable approximation during the backward pass of training. This enables the use of gradient-based learning algorithms such as backpropagation in the training process of SNNs.

Despite notable successes in training spiking neural networks, backpropagation has faced persistent criticism for its training challenges and lack of alignment with the mechanisms of the brain~\cite{bpbrain1}, raising doubts about its viability as a model for credit assignment in neural systems~\cite{bpandbrain}. 
One of the major limitations lies in its requirement to explicitly store all neural activity for later use during synaptic adjustments. This explicit storage mechanism has no direct counterpart in biological neural circuits, where memory and computation are more integrated~\cite{bpweakness}.
Another significant challenge is the requirement for error derivatives to propagate through a dedicated global feedback pathway within the network to generate teaching signals. This kind of structured, system-wide error communication contrasts sharply with the distributed and localized nature of information processing in the brain~\cite{bpweakness2}. It may, in certain cases, exacerbate issues such as vanishing and exploding gradients in algorithmic implementations.
Furthermore, backpropagation assumes that global error signals can move backward through the same neural pathways used for forward propagation, a phenomenon referred to as the weight transport problem~\cite{wtproblem}. This assumption is biologically implausible, as neural pathways in the brain are not designed to facilitate such exact bidirectional transport~\cite{wtproblem2}.
The sequential nature of backpropagation adds another layer of incompatibility. Unlike the brain’s ability to perform massively parallel computations, backpropagation enforces a process where inference and learning must occur largely step-by-step. This sequentiality significantly diverges from biological neural networks' highly dynamic and concurrent operations~\cite{bpparallel}. 
These challenges highlight why backpropagation is unlikely to serve as a biologically plausible brain learning model. However, recent advancements have been exploring alternative approaches~\cite{b21} that address some of these limitations and offer solutions more in line with biological principles.

The Forward-Forward algorithm~\cite{hinton2022}, inspired by contrastive methods like Boltzmann Machines~\cite{boltzman} and Noise Contrastive Estimation~\cite{ncl}, is a novel alternative to the traditional backpropagation method for training neural networks, which Geoffrey Hinton introduced. This algorithm replaces backpropagation's forward and backward passes with two forward passes. Each network layer has its own objective function to optimize the corresponding layer's weights. FF does not require storing all intermediate activities for updating each layer's weights, nor does it assume symmetrical weight connections. This makes it a more plausible model for cortical learning and suitable for low-power analog hardware~\cite{hinton2022}. Also, FF can process data sequentially, enabling on-the-fly learning without halting the data pipeline for all the weights' updating. Another feature of FF is that, unlike backpropagation, it does not require direct connectivity of all consecutive layers through the forward pass, allowing for the inclusion of black-box components within the network.

Motivated by the advantages of the FF algorithm and its potential for SNN training, this work develops an FF-based SNN learning approach and investigates its performance on static and spiking benchmark datasets. We summarize our contributions as follows:

\begin{itemize}
    \item A novel gradient-based learning approach is introduced for training spiking neural networks inspired by the Forward-Forward algorithm, which demonstrates strong performance across a range of benchmark datasets.
    
    \item A structured training approach for spiking neural networks that leverages key principles of the Forward-Forward algorithm, including layer-wise local learning, black-box compatibility, and additional advantageous features.

    \item Experimental results of our SNN model, trained using the FF algorithm on spiking datasets such as N-MNIST and SHD, demonstrate competitive performance compared to backpropagation-based models and establish the FF approach as a viable alternative for training spiking neural networks.
   
\end{itemize}

The rest of this paper is organized as follows: Section~\ref{sec2} overviews related work and backgrounds of SNNs and the FF algorithm. Section~\ref{sec3}  presents the proposed method. Section~\ref{sec4}  elaborates on the experimental setup and results. Finally, Section~\ref{sec5} concludes the paper and suggests a direction for future work.

\begin{figure*}[t]
\centering
\includegraphics[width=0.9\textwidth]{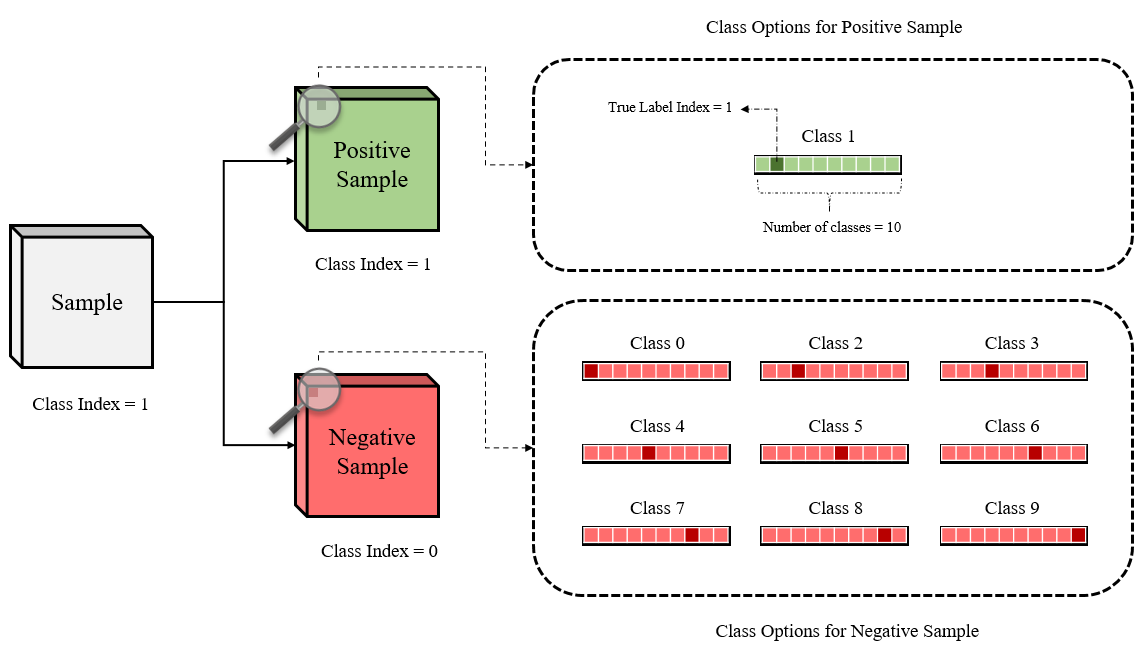}
\caption{Different options for overlaying a label on a sample of a 10-class dataset for positive and negative sample generation.}
\label{fig:label}
\end{figure*}
%%%%%%%%%%%%%%%%%%%%%%%%%%%%%%%%%%%%%%%%%%%%%%%%%%%%%%%%%%%%%%%%%%%%%%%%%%%%%%%%%%%%%%%%%%%%%%%%%%%%

\section{Related Work and Background}
\label{sec2}
This section provides an overview of the literature relevant to our work, covering learning approaches for spiking neural networks, the Forward-Forward algorithm, and its application within the context of SNNs.

\subsection{Learning Approaches for SNNs}
Since the emergence of spiking neural networks, various ways have been used to train them. These learning approaches are generally divided into three categories: STDP, ANN-SNN conversion, and Backpropagation.
STDP~\cite{stdp} is an unsupervised learning algorithm inspired by biological neural systems, which adjusts synaptic weights based on the relative timing of presynaptic and postsynaptic spikes. The fundamental principle is that the order of these spikes determines whether the synapse is strengthened or weakened. In Hebbian STDP, a presynaptic spike occurring before a postsynaptic spike leads to synaptic strengthening, reinforcing the association between neurons. Conversely, in anti-Hebbian STDP, the same spike timing results in synaptic weakening, promoting different forms of neural plasticity~\cite{hebbian}. Some variants of the STDP approach include Reward-modulated STDP~\cite{rstdp}, Mirrored STDP~\cite{mstdp}, and Probabilistic STDP~\cite{pstdp}, each of which can be utilized for different purposes.

ANN-to-SNN conversion approach is a widely used method for leveraging the strengths of both networks~\cite{ann2snn}. It involves training a standard ANN using traditional methods and then adapting it into an SNN by mapping its weights and neuron parameters to spiking equivalents. This approach combines the efficiency of ANN training with the benefits of SNN operation, allowing SNNs to achieve performance comparable to ANNs on some benchmarks.

Backpropagation in SNNs focused on understanding and modeling complex spatiotemporal relationships between spikes. This method prioritizes computational effectiveness rather than adhering to biological plausibility, making it distinct from approaches like STDP. A primary challenge with backpropagation is the nondifferentiable nature of spiking neuron equations, which prevents the direct application of standard gradient-based optimization methods. Various approximation techniques like SpikeProp~\cite{SpikeProp}, SuperSpike~\cite{superspike}, SSTDP~\cite{sstdp}, and SLAYER~\cite{slayer} are employed to overcome this, enabling backpropagation to function effectively in the spiking domain. 

\subsection{Forward-Forward Algorithm}
The Forward-Forward Algorithm is a novel learning approach for neural networks that forgoes traditional backpropagation by using two forward passes: one with real or "positive" data and another with "negative" or contrasting data. First, the algorithm prepares data by creating positive samples, which are real labeled data, and generating negative samples. These negative samples are carefully crafted, either by corrupting positive samples or using synthetic data, to differ structurally from positive data. In supervised learning, in order to use labels in the training process, positive and negative samples can be generated from the original data in various ways. One common way is to embed the corresponding labels on each sample. When embedding, it should be noted that if we are creating positive samples, we must put the correct label on the data, and if we are looking to create negative samples, we must put the wrong label on the data. 

Figure~\ref{fig:label} shows how to embed labels into an image from a static dataset and create positive and negative samples corresponding to a sample from class one from a dataset that has 10 classes. 
It can be seen that in the positive sample corresponding to the selected sample from class one, the second pixel is lit out of the first ten pixels. Similarly, in general, in creating its corresponding negative sample, any pixel except the second pixel can be lit out of the first ten pixels, which, for example, class zero is considered as the class of the negative sample.

The FF algorithm then processes each sample in two separate forward passes through the network, each with distinct objectives.
In the first forward pass, positive data is fed through the network, and the network’s goal is to maximize a metric called “goodness” in each layer. Goodness typically represents the sum of squared neuron activations within a layer, measuring that input's “energy” or response level. During the second forward pass, the same network processes the negative samples, but this time it seeks to minimize the goodness measure, reducing neuron activations in response to negative samples. Each layer thus learns independently by adjusting its weights based on local goodness functions, making the FF algorithm highly adaptable without needing backward error propagation.

To ensure the network learns robust representations, FF applies layer normalization. This step forces layers to focus on the orientation of activations rather than just their magnitude, preventing lower layers from overpowering the classification and encouraging meaningful feature extraction across all layers. In a deep network, this allows each layer to contribute to classification rather than relying solely on early layers.
Training proceeds sequentially across layers, following a greedy layer-wise approach that simplifies learning since each layer has its objective and does not depend on global backpropagation. For different tasks, FF adapts to both unsupervised and supervised learning. Unsupervised learning uses synthetic negative samples to differentiate representations, while in supervised learning, correct labels in the input indicate positive data, and incorrect labels create negative samples. 
The FF algorithm has practical advantages, notably eliminating the need for backpropagation and reducing computational demands~\cite{danilo}. This makes it promising for online learning and analog hardware applications where backpropagation is challenging~\cite{hinton2022}. 

\subsection{Forward-Forward Algorithm in SNNs}
Despite the widespread use of the BP method in learning spiking neural networks, problems such as weight transport, update locking, and black-box handling still persist. Using the Forward-Forward algorithm instead of backpropagation and replacing its steps in the training process of spiking neural networks can solve these problems to a good extent. Studies have also been conducted using the FF algorithm's learning idea to spiking neural networks. 

Ororbia~\cite{csdp2024} propose a biologically plausible learning framework for spiking neural networks called Contrastive Signal–Dependent Plasticity (CSDP). This method enables self-supervised learning by combining local plasticity rules with contrastive principles derived from forward-forward learning. The architecture consists of recurrent layers of leaky integrate-and-fire neurons that are connected via top-down, bottom-up, and lateral synapses. These layers operate in parallel and asynchronously, allowing for efficient and scalable simulation while avoiding biologically implausible constraints such as feedback connections or backpropagation through time.
At the core of the learning process is an activation trace that smooths sparse spike activity over time. Each layer adapts its synapses based on a local contrastive objective that encourages the network to increase the “goodness” or likelihood of in-distribution (positive) inputs and suppress the response to out-of-distribution (negative) inputs. This is achieved using a Hebbian-like synaptic update rule modulated by a local contrastive signal, eliminating the need for non-local information such as global error signals or explicit error feedback pathways.
Negative samples required for contrastive learning are generated dynamically. In the supervised setting, they are created by randomly assigning incorrect labels to cloned inputs. In the unsupervised case, they are synthesized using random augmentations and interpolations of input patterns within a batch. The model is further extended to perform task-specific functions such as classification and reconstruction by adding simple generative and predictive synaptic pathways that follow similar local learning rules.

Terres et al.~\cite{snnffrobust} introduces a framework for robust Out-of-Distribution (OoD) detection using fully-spiking neural networks trained with the Forward-Forward algorithm, a forward-only alternative to backpropagation. They adapt FF to the spiking domain by proposing two spiking-compatible goodness functions. One is unbounded to mimic ReLU behavior via spike aggregation, and one is bounded, aligning with the discrete, time-limited spiking activity.
They then propose FF-SCP, a new distance-based OoD detection algorithm that uses the latent space structure induced by FF. Instead of relying on class centroids, FF-SCP calculates the minimum distance between a sample's latent representation and class-specific latent manifolds, providing a more accurate reflection of class boundaries. This distance serves as an OoD score.
To improve explainability, the authors further design a gradient-free attribution technique that uses a decoder to reconstruct inputs from latent vectors. By optimizing these latent representations to match a target class distribution while staying close to the original input, the method generates attribution maps that highlight features responsible for OoD classification.
Experiments across datasets such as MNIST, F-MNIST, K-MNIST, and Not-MNIST show that the proposed spiking FF system achieves OoD detection performance comparable to or better than state-of-the-art spiking and non-spiking methods like SCP and ODIN, with strong energy-efficiency implications due to the spiking architecture.
%%%%%%%%%%%%%%%%%%%%%%%%%%%%%%%%%%%%%%%%%%%%%% Proposed Method %%%%%%%%%%%%%%%%%%%%%%%%%%%%%%%%%%%%%%%%%%%%%%%%%%%
\section {Proposed Method}\label{sec3}
The spiking neural networks training procedure using the Forward-Forward algorithm involves generating labeled training samples that align with the algorithm's layer-wise contrastive learning framework. A critical step in this process is superimposing class labels onto the input vectors by replacing part of the original input with class-specific information. This enables the network to evaluate whether a given input-label pair is compatible, allowing each layer to independently distinguish between positive (correct) and negative (incorrect) associations.
Consider a flattened input vector $x^{input} \in \mathbb{R}^{d}$ like:
\begin{equation}
\begin{aligned}
x^{input} = [x_{0}, x_{1}, \dots, x_{d-1}].
\end{aligned}
\end{equation}

To create these labeled inputs from both static and spatiotemporal data, we modify the input vector by replacing the first $c$ elements of the flattened vector, where $c$ is the number of classes, with a label-encoding vector. Specifically, we initialize these $c$ elements to zero and assign the maximum element's value of the original input, denoted as $m$, to the index corresponding to the label class:
\begin{equation}
\begin{aligned}
m = \max_{0 \leq i \leq d-1} x^{\text{input}}_i ,  
\end{aligned}
\end{equation}
the remaining $ d-c$ elements retain the corresponding values from the original input vector $x^{input}$. The modified vector ${x^{pos}} \in \mathbb{R}^{d}$ for the positive sample with true label $y^{pos} = t$ is constructed as:
\begin{equation}
\begin{aligned}
x_{i}^{pos} = 
\begin{cases}
      0 & i\leq c \ and \ i\neq t , \\
      m & i=t , \\
      x_{i} & i \geq c ,
\end{cases}
\end{aligned}
\label{x_pos}
\end{equation}
where the value $m$ replaces the $t^{th}$ entry (index $t-1$) to indicate the true label, while the remaining elements $x_{c}$ to $x_{d-1}$ are directly copied from the original input.
To construct a negative sample, we choose a false label $y^{\text{neg}} = f \neq t$ , and embed it as:
\begin{equation}
\begin{aligned}
x_{i}^{neg} = 
\begin{cases}
      0 & i\leq c \ and \ i\neq f , \\
      m & i=f, \\
      x_{i} & i \geq c .  
\end{cases}
\end{aligned}
\label{x_neg}
\end{equation}
\begin{figure*}[t]
\centering
\includegraphics[width=0.97\textwidth]{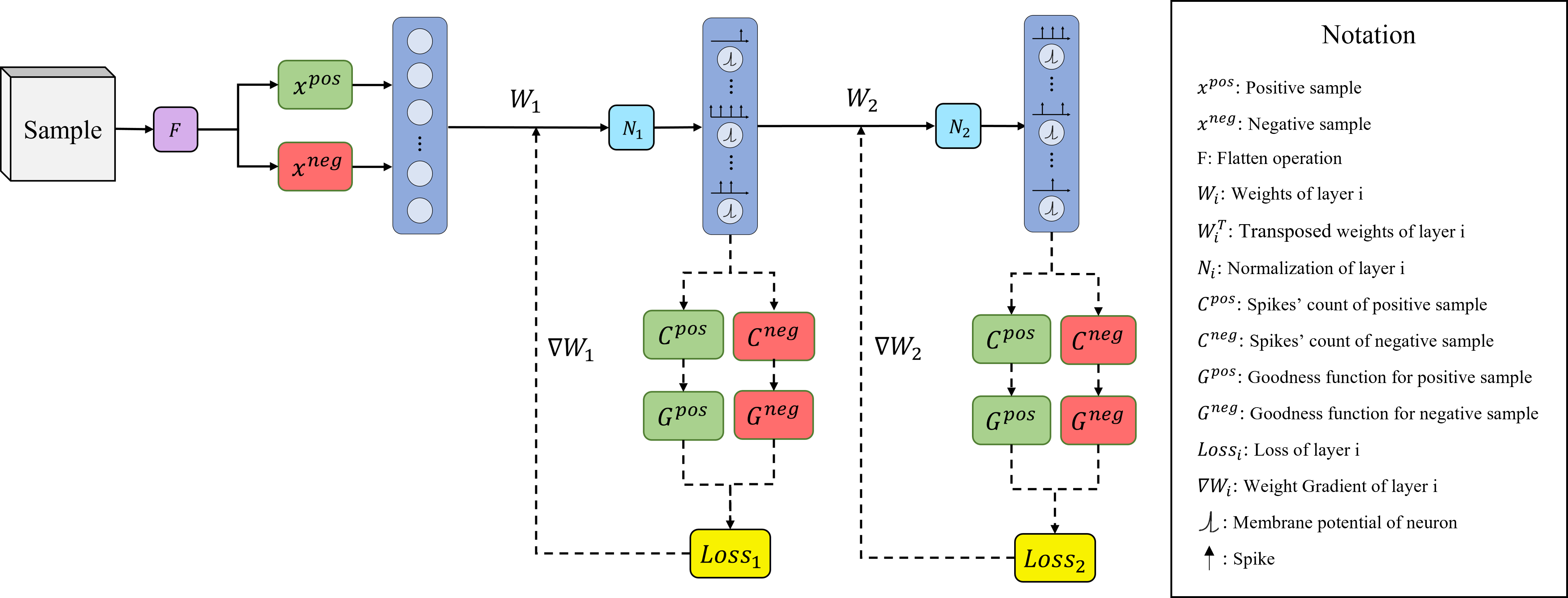}
\caption{The architecture of the proposed spiking forward-forward algorithm. The diagram highlights the input forward path (solid arrows) and the error propagation path (dotted arrows).}
\label{fig:model}
\end{figure*}
In this case, the embedded label does not match the ground truth, and the network is expected to learn to distinguish positive samples from negative samples.
This label replacement strategy embeds class identity directly into the input vector without changing its dimensionality. It supports forward-only contrastive training by enabling each layer to compute a scalar goodness score and adjust weights locally, eliminating the need for explicit output layers or backpropagated error signals.

To choose the index $f$ for negative samples more intelligently instead of randomly, we use hard labeling to assign the labels of negative samples.  For each input image, the model computes a "goodness" score for all possible class labels. The goodness score measures the mean squared activation of the network layers in response to the input. The goodness score for the true class label of each input image is set to zero. This ensures that the true label is not considered when selecting a hard label. The remaining goodness scores (excluding the true label) are transformed using a square root function. This transformation flattens the distribution, making it less peaked and more uniform, thus avoiding the dominance of any single class label. Using the transformed goodness scores as a probability distribution, a hard label is sampled for each input image. This label is chosen from the set of non-true labels and represents a class that the network finds relatively difficult to distinguish from the true label. Hard labeling aims to create challenging negative examples for the network. 
After generating the positive and negative samples, they are fed into the network, which consists of an input layer followed by fully connected hidden layers. Notably, the network does not include an output layer, and the number of hidden layers can be adjusted based on the complexity of the task. Following a linear transformation of the input data, a specified neuron model—such as the leaky integrate-and-fire (LIF) model—is applied. 
LIF neurons accumulate input current over time and emit a spike when the membrane potential exceeds a predefined threshold. The membrane potential then resets, mimicking the functions observed in biological neurons. The neurons are initialized with parameters like the leakage term and the spiking threshold. The recursive representation of a LIF neuron is as follows:
\begin{equation}
\begin{aligned}
U[t+1] = \beta U[t] + WX[t+1]- R[t],
\end{aligned}
\end{equation}
where $U$, $W$, $X$, $R$, and $\beta$ are membrane potential, weights, inputs, reset term, and decay rate, respectively. 
In each time step, spike emission occurs if the membrane potential exceeds the threshold $U^{thr}$:
\begin{equation}
\begin{aligned}
S[t] = 
\begin{cases}
      1 & U^{thr}\leq U[t], \\
      0 & otherwise .  
\end{cases}
\end{aligned}
\label{spike function}
\end{equation}

%###############
The Forward-Forward algorithm aims to increase neuronal activation in response to positive samples while suppressing activation for negative samples in every layer. However, directly propagating neuron outputs to the next hidden layer can impede effective learning in deeper layers, especially for positive samples. This challenge stems from the accumulation of large activations from preceding layers, which can dominate the learning signal. To address this, a normalization step~\cite{BN} is applied to the output of each layer's linear transformation (i.e., the input to the subsequent layer) at every time step, allowing each hidden layer to learn independently of its predecessors. In each layer, the normalization performed at every time step \( t \) is defined as follows:
\begin{equation}
\begin{aligned}
{N}(WX[t]) = \gamma [t] \odot \frac{WX[t] - \mu[t]}{\sqrt{\sigma^2 [t] + \epsilon}} + \beta [t], 
\end{aligned}
\end{equation}
where the parameters $\gamma$ and $\beta$ are learned during training. The constant $\epsilon$ is a small value added to the variance to prevent division by zero during normalization, and the symbol $\odot$ represents the Hadamard product, an element-wise multiplication operation. \( \mu \) and \( \sigma^2 \) are the mean and variance along the samples of the mini-batch, respectively, and for each layer, they are defined as follows:
\begin{equation}
\begin{aligned}
\mu [t] &= \frac{1}{B} \sum_{b=1}^{B} WX_{b} [t] , \\
\sigma ^2 [t] &= \frac{1}{B} \sum_{b=1}^{B} \left(WX_{b} [t] - \mu [t]\right)^2 ,
\end{aligned}
\end{equation}
where $B$ is the batch size. Accordingly, the formulation of the LIF neuron model can be reformulated as:
\begin{equation}
\begin{aligned}
U[t+1] = \beta U[t] + N(WX[t+1]) - R[t].
\end{aligned}
\end{equation}
%################

Our method involves two distinct forward passes, the positive and negative passes. During the positive pass, positive samples are presented to the network, and at each time step, the spiking condition for each neuron is evaluated. The goodness value for each hidden layer is then computed by applying a goodness function to the spike count of each neuron. Subsequently, the negative pass is performed, where negative samples are propagated through the network using the same procedure to compute their corresponding goodness values. Throughout both the positive and negative forward passes, the spike count for each neuron $i$, defined as the total number of spikes accumulated over all time steps, is computed as follows:
\begin{equation}
    C_{i}^{pos} = \sum\limits_{t=1}^T S_{i}^{pos}[t] ,  \\
\end{equation}
\begin{equation}
    C_{i}^{neg} = \sum\limits_{t=1}^T S_{i}^{neg}[t] ,
\end{equation}
where $T$ denotes the total number of time steps, while \( S_{i}^{\text{pos}} \) and \( S_{i}^{\text{neg}} \) represent the spike occurrences of neuron $i$ in response to positive and negative samples, respectively. In each layer, the goodness function \( G \) for positive and negative samples is defined as follows:
\begin{equation}
    G^{pos} = \frac{1}{N}\sum\limits_{i=1}^N (C^{pos}_{i})^2 ,  \\
\end{equation}
\begin{equation}
    G^{neg} = \frac{1}{N}\sum\limits_{i=1}^N (C^{neg}_{i})^2 ,  \\
\end{equation}
where $N$ is the number of neurons in the given layer. Based on the goodness values computed for positive and negative samples, the loss for each layer is calculated independently using a function inspired by the Swish function~\cite{swish}, defined as follows:
\begin{equation}
\begin{aligned}
&\Delta = G^{\text{pos}} - G^{\text{neg}}  ,\\
&Loss = \frac{-\alpha\Delta}{1 + e^{\alpha\Delta}} ,
\end{aligned}
\end{equation}
\begin{figure}[t]
    \centering
    \includegraphics[width=0.45\textwidth]{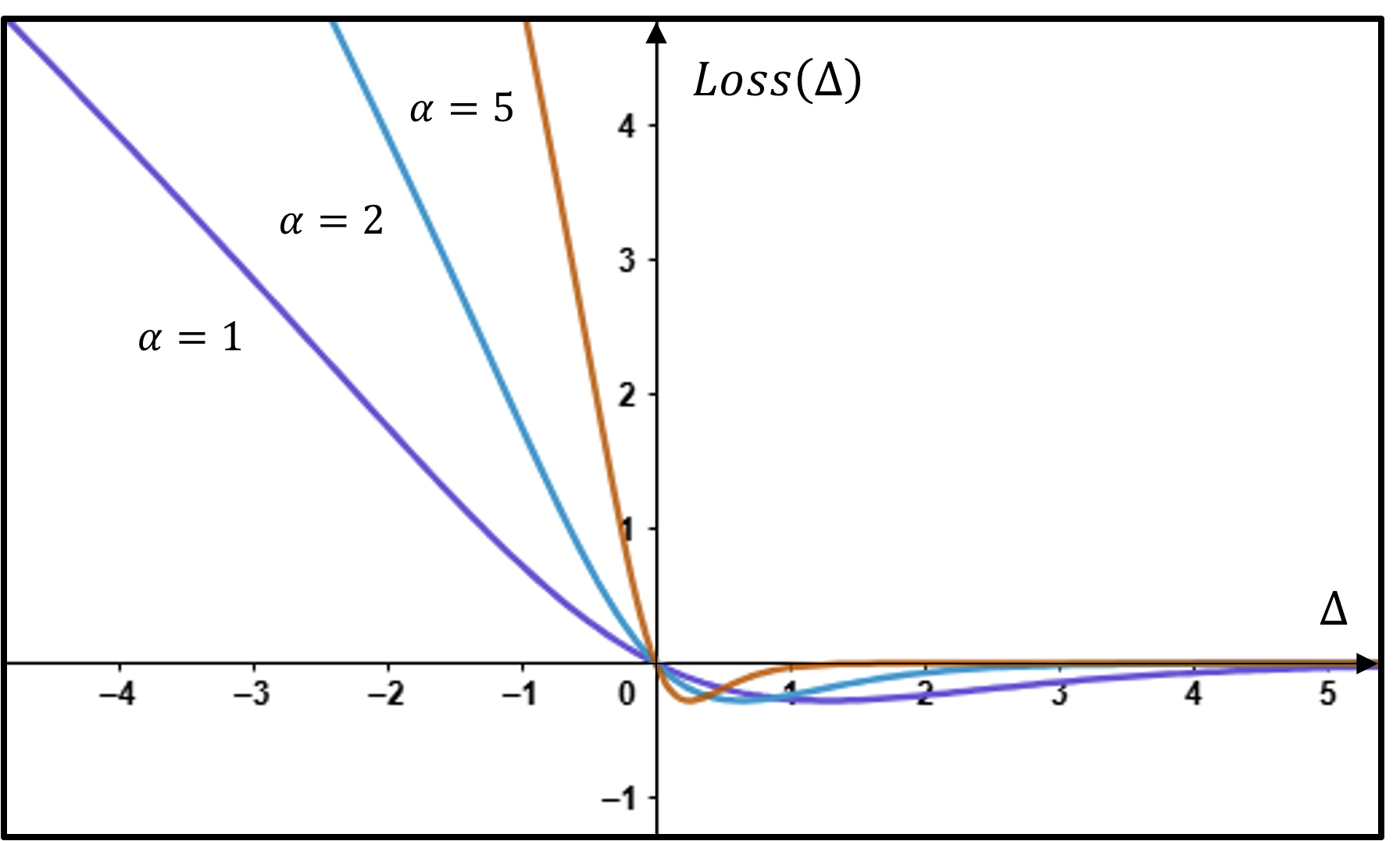}
    \caption{Loss function $\frac{-\alpha\Delta}{1 + e^{\alpha\Delta}}$ plotted against $\Delta$ for different scaling factor values of $\alpha$.}
    \label{fig:loss_graph}
\end{figure}
where $\alpha$ is a scaling factor that controls the sharpness of the loss curve. Figure~\ref{fig:loss_graph} illustrates the behavior of the loss function plotted against the variable $\Delta$ for different values of the scaling factor $\alpha \in \{1, 2, 5\}$. 
This function serves as a smooth contrastive objective, designed to increase the margin between positive and negative goodness values. As depicted, for negative values of $\Delta$, the loss increases approximately linearly, with steeper slopes for larger values of $\alpha$. This implies stronger penalization when the goodness of negative samples surpasses that of positive samples. In contrast, for the large positive values of $\Delta$, the loss asymptotically approaches zero, indicating that well-separated positive and negative samples contribute less to the gradient, thereby preventing excessive updates once correct classification with sufficient margin has been achieved. Higher values of $\alpha$ result in sharper transitions near $\Delta = 0$, effectively tightening the decision boundary and making the loss function more sensitive to misclassified or marginal examples. This makes $\alpha$ a critical hyperparameter for modulating margin sensitivity and learning dynamics. The weights of each layer are updated to minimize the loss value by computing the gradient of the loss with respect to the layer's weights.
Training of spiking neural networks is challenging due to the non-differentiability of the spike generation process. To overcome this, surrogate gradient learning~\cite{neftci2019} can be utilized. Surrogate gradients provide a differentiable approximation of the spike function, allowing gradient-based optimization methods to be used. During layer-wise weight updating, these surrogate gradients approximate the true gradients of the spike function, enabling the network to update its parameters effectively. Guided by experimental results, we utilize the shifted arctangent surrogate function \cite{ann2snn}, with its gradient with respect to the membrane potential expressed as:
\begin{equation}
\begin{aligned}
\label{spike_occurance}
S[t] &\approx \frac{1}{\pi}arctan(\pi \frac{\alpha}{2}) U[t] ,\\
\frac{\partial S[t]}{\partial U[t]} &= \frac{1}{\pi}\frac{1}{(1+(\pi \frac{\alpha}{2} U[t])^{2})},
\end{aligned}
\end{equation}
where $\alpha$ is the slope parameter. After updating the weights of each layer, the next epoch of training is implemented.
Figure~\ref{fig:model} illustrates the learning procedure of the proposed method. It is assumed that the training operation is to be performed on an original sample. By applying the label embedding operation, the sample is converted to a positive and negative sample, and before entering the fully connected neural network, each positive and negative sample must be flattened and converted to a vector. In the positive forward pass, the positive sample enters the neural network, and spikes are generated in the neurons during the time steps. In each neuron, the number of spikes is counted independently, and after that, their spike count passes to the goodness function $G$ to calculate the goodness value of the positive sample. For the negative sample, the same steps are followed, and at the end, the goodness of the negative sample is calculated. After finding the goodness values for positive and negative samples, the amount of loss is calculated with the loss function $L$, and then the weights of the corresponding layer are updated locally.

During inference, the absence of a dedicated output layer precludes the use of conventional classification methods. To address this, we adopt a label-scoring strategy that leverages layer-wise information for prediction. Specifically, we iterate over all possible class labels for each test sample by embedding each label into the input, as done during training. Each resulting input-label pair is passed through the trained network, and the goodness values computed across all layers are summed. The label that yields the highest total goodness score is then selected as the predicted class for the sample. 
%%%%%%%%%%%%%%%%%%%%%%%%%%%%%%%%%%%%%%%%%%%%%%%%%%%%%%%%%%%%%%%%
\begin{table*}[t]\label{table:hyperparameters}
\caption{Datasets and corresponding training parameters.}
\label{table:hyperparams}
\begin{center}
\resizebox{0.8\textwidth}{!}{
\renewcommand{\arraystretch}{1.3}
\begin{tabular}{cccccccccc}
\hline
&\multicolumn{1}{c}{\bf MNIST} &\multicolumn{1}{c}{\bf F-MNIST} &\multicolumn{1}{c}{\bf K-MNIST} &\multicolumn{1}{c}{\bf CIFAR-10} &\multicolumn{1}{c}{\bf N-MNIST} &\multicolumn{1}{c}{\bf SHD}\\
\hline
\multicolumn{1}{c}{\multirow{1}{*}{Dataset (train/test)}} &$60\text{k}/10\text{k}$ &$60\text{k}/10\text{k}$ &$60\text{k}/10\text{k}$ &$50\text{k}/10\text{k}$  &$60\text{k}/10\text{k}$&$8156/2264$ \\
\multicolumn{1}{c}{\multirow{1}{*}{Input neurons}} &$784$ &$784$ &$784$ &3072 &$2312$ &$700$ \\
\multicolumn{1}{c}{\multirow{1}{*}{Dataset classes}} &$10$ &$10$ &$10$ &$10$ &$10$ &$20$ \\
\multicolumn{1}{c}{\multirow{1}{*}{Epochs}} &$300$ &$300$ &$300$ &$300$ &$300$ &$500$ \\
\multicolumn{1}{c}{\multirow{1}{*}{Learning rate}} &$0.001$ &$0.001$ &$0.001$ &$0.001$ &$0.001$ &$0.001$  \\
\multicolumn{1}{c}{\multirow{1}{*}{Batch size}} &$4096$ &$4096$ &$4096$ &$4096$ &$4096$ &$4096$  \\
\multicolumn{1}{c}{\multirow{1}{*}{Time steps $T$}} &$10$ &$10$ &$10$ &$10$ &$10$ &$10$ \\
\multicolumn{1}{c}{\multirow{1}{*}{$\alpha$ (in loss)}} &$0.6$ &$0.6$ &$0.6$ &$0.6$ &$0.6$ &$0.6$ \\
\multicolumn{1}{c}{\multirow{1}{*}{Neurons' threshold}} &$1.0$ &$1.0$ &$1.2$ &$1.2$ &$1.0$ &$5.0$ \\
\multicolumn{1}{c}{\multirow{1}{*}{Neurons' decay rate}} &$0.99$ &$0.99$ &$0.99$ &$0.8$ &$0.9$ &$0.9$ \\
\hline
\end{tabular}
}
\end{center}
\end{table*}
%%%%%%%%%%%%%%%%%%%%%%%%%%%%%%%%%%%%%%%%%%
%--------------------------------Experoments---------------------------
\section{Experiments}
\label{sec4}
To evaluate the effectiveness of the proposed approach, we conducted a series of experiments. This section is organized into two parts. First, the \textit{Experimental Setup} outlines our study's datasets and implementation details. Then, the \textit{Experimental Results and Analysis} section reports the performance outcomes, compares with baseline approaches, and discusses key findings and insights derived from the results.
\subsection{Experimental Setup}
The experiments were conducted using a PyTorch-based implementation~\cite{pytorch}, leveraging the snntorch library~\cite{jason2023} for spiking neuron models and surrogate gradients. To evaluate the proposed method, six datasets were utilized: MNIST~\cite{MNIST}, Fashion-MNIST (F-MNIST)~\cite{FMNIST}, Kuzushiji-MNIST (K-MNIST)~\cite{KMNIST}, CIFAR-10~\cite{CIFAR10}, Neuromorphic-MNIST (N-MNIST)~\cite{NMNIST}, and Spiking Heidelberg Digits (SHD)~\cite{cramerSHD}.
The MNIST dataset comprises 70,000 grayscale images of hand-written digits, each 28×28 pixels wide. It is divided into 60,000 samples for training and 10,000 for testing, covering digit classes from 0 to 9. 
F-MNIST and K-MNIST are both drop-in replacements for the original MNIST dataset. They offer more challenging classification tasks while maintaining the same 28×28 grayscale image format and 10-class structure. F-MNIST consists of images of clothing items such as shirts, shoes, and bags, aiming to test visual recognition models beyond simple digits. K-MNIST, on the other hand, features cursive Japanese kana characters derived from historical texts, introducing additional complexity due to the intricate and variable nature of the script. 
The CIFAR-10 dataset includes 60,000 color images, each sized 32×32 pixels with three color channels. It spans 10 object classes, with 50,000 samples allocated for training and 10,000 for testing. Unlike MNIST and Fashion-MNIST, which use grayscale images, CIFAR-10 focuses on more visually intricate static 2D color images.
The N-MNIST dataset is derived from the MNIST dataset by applying a spiking conversion. This process involves recording digit samples displayed on an LCD monitor using an ATIS sensor mounted on a motorized pan-tilt unit, which introduces motion-based neuromorphic features while retaining the same classes and sample count as MNIST.
Lastly, the SHD dataset is used, which includes high-quality studio recordings of spoken digits in both German and English. The original dataset contains 10,420 samples, which are divided into 8,156 training and 2,264 testing samples spanning 20 classes.

The network parameters are initialized, and the Adam optimizer is configured with an initial learning rate of 0.001. The training is conducted for 300 epochs, with adjustments to the learning rate based on epoch milestones to fine-tune the learning process. The training data is split into mini-batches of 4096 samples each. For each mini-batch, positive and negative examples are generated by superimposing true and hard labels onto the input vectors. The hyperparameter values used to train the model on each dataset are summarized in Table~\ref{table:hyperparams}.
%-------------------------------------
\subsection{Experimental Results and Analysis}
Our experimental results and comparisons are presented in two separate tables. Table~\ref{table:results} provides a comparison between our FF-based SNN model and other existing SNN models, while Table~\ref{table:results2} focuses on comparing our SNN model with various FF-based ANN implementations.
Table~\ref{table:results} comprehensively compares spiking neural network models across five datasets: MNIST, F-MNIST, K-MNIST, N-MNIST, and SHD. The key columns detail the model name, architecture, neuron model type, and corresponding accuracy percentage, highlighting both our proposed model and various state-of-the-art models with a fully connected architecture. 
%%%%%%%%%%%%%%%%%%%%%%%%%%%%%%%%%%%%%%%%%%

In the MNIST dataset, our model achieves an impressive 98.34\% accuracy using a LIF$^{\beta}$\cite{PLIF} neuron model and a two-hidden-layer architecture with 500 neurons, outperforming other models. 
Compared to the FF-based models, Ororbia et al.~\cite{csdp2024} use a deeper model with two hidden layers of 7000 neurons, achieving 97.46\% with LIF neurons, and Terres et al.~\cite{snnffrobust} achieve lower accuracy with 93.27\%.
Competing BP-based models from Neftci et al.~\cite{neftci2019}, Comsa et al.~\cite{comsa2020}, and Taylor et al.~\cite{oxford2023} use slightly different architectures and neuron models like IF or LIF$^{\beta}$, achieving comparable accuracies 97.87\%, 97.90\%, and 97.91\%, respectively. However, our model outperforms these models, showcasing an improvement, potentially due to the specific architecture and neuron model used.
\begin{table*}[t]
\caption{Performance of our SNN model compared to the existing SNN models (* denotes self-implementation, $^\dagger$ denotes data augmentation, and $^\beta$ denotes learnable membrane time constants).}
\label{table:results}
\begin{center}
\resizebox{\textwidth}{!}{
\renewcommand{\arraystretch}{1.3}
\begin{tabular}{cccccccccc}
\hline
\multicolumn{1}{c}{\bf Dataset}  &\multicolumn{1}{c}{\bf Model} &\multicolumn{1}{c}{\bf Architecture} &\multicolumn{1}{c}{\bf Neuron model} &\multicolumn{1}{c}{\bf Learning Approach} &\multicolumn{1}{c}{\bf Accuracy (\%)}  \\
\hline

% MNIST
\multicolumn{1}{c}{\multirow{2}{*}{MNIST}}
	& \textbf{Proposed model}  &784-500-500 &LIF$^\beta$ &FF  &$98.34$ \\
	& Ororbia.~\cite{csdp2024}  &784-7000-7000 &LIF &FF  &97.46 \\
	& Terres et al.~\cite{snnffrobust}  &784-1400-1400 &LIF &FF &93.27 \\
	& Taylor et al.~\cite{oxford2023} $^\dagger$  &784-1000-10 &LIF$^\beta$ &BP  &97.91 \\
	& Comsa et al.~\cite{comsa2020}  &784-340-10 &IF (alpha-PSP) &BP &97.90 \\
	& Neftci et al.~\cite{neftci2019}$^\dagger$  &784-1000-10 &LIF$^\beta$ &BP &97.87 \\
	
    \hline
   
% F-MNIST
\multicolumn{1}{c}{\multirow{2}{*}{F-MNIST}}
    &\textbf{Proposed model}  &784-500-500 &LIF$^\beta$ &FF &89.36 \\
    & Terres et al.~\cite{snnffrobust}  &784-1400-1400 &LIF &FF &85.68 \\
	& Neftci et al.~\cite{neftci2019}$^\dagger$  &784-1000-10 &LIF$^\beta$ &BP &89.93 \\
	& Taylor et al.~\cite{oxford2023}$^\dagger$  &784-1000-10 &LIF$^\beta$ &BP &89.05 \\
	& Zhang et al.~\cite{zhang2022}$^\dagger$  &784-1000-10 &IF(ReL-PSP) &BP &88.1 \\
	& Kheradpisheh et al.~\cite{kheradpisheh2020}$^\dagger$  &784-1000-10 &IF &BP &88.0 \\
    & Perez-Nieves et al.~\cite{sparse2021}  &784-200-10 &LIF &BP &82.20 \\
    
    \hline

%K-MNIST
   \multicolumn{1}{c}{\multirow{2}{*}{K-MNIST}}
   &\multicolumn{1}{c}{\multirow{1}{*}{\textbf{Proposed model}}}  &784-500-500 &LIF$^\beta$ &FF &91.26 \\
   & Ororbia.~\cite{csdp2024}  &784-7000-7000 &LIF &FF  &91.14 \\
	& Terres et al.~\cite{snnffrobust}  &784-1400-1400 &LIF &FF &85.91 \\
   
   \hline

% N-MNIST
\multicolumn{1}{c}{\multirow{2}{*}{N-MNIST}}
&\multicolumn{1}{c}{\multirow{1}{*}{\textbf{Proposed model}}}  &2312-500-500 &LIF$^\beta$ &FF &97.26 \\
	  & Neftci et al.~\cite{neftci2019}  &2312-300-10 &LIF$^\beta$ &BP &97.47 \\
        & Taylor et al.~\cite{oxford2023} &2312-300-10 &LIF &BP &95.91 \\
        & Perez-Nieves et al.~\cite{sparse2021}  &2312-200-10 &LIF &BP &92.70 \\
\hline

% SHD
\multicolumn{1}{c}{\multirow{2}{*}{SHD}}
        & \textbf{Proposed model}  &700-500-500 &recurrent LIF &FF &77.87 \\
        & Cramer et al.~\cite{cramerSHD}  &700-128-20 &recurrent LIF &BP &71.40 \\
        & Neftci et al.~\cite{neftci2019}  &700-300-20 &LIF$^\beta$ &BP &70.81\\
        & Cramer et al.~\cite{cramerSHD}  &700-128-20 &LIF &BP &48.1 \\

\hline
\end{tabular}}
\end{center}
\end{table*}

In the F-MNIST, our model achieves an accuracy of 89.36\% with the same two hidden layers of 500 neurons architecture and LIF$^{\beta}$ neuron model. Compared to an FF-based model, Terres et al.~\cite{snnffrobust} use two hidden layers of 1400 neurons, achieving 85.68\%, which our model achieves higher accuracy despite a lighter network.
Compared to BP-based models, Zhang et al.~\cite{zhang2022} and Kheradpisheh et al.~\cite{kheradpisheh2020} report slightly lower accuracies (88.1\% and 88.0\%) using IF-type neuron models with a 1000-neuron hidden layer architecture. The other competing models, Taylor et al.~\cite{oxford2023} achieve closer performance to our model employing LIF$^{\beta}$ with 89.05\% and Neftci et al.~\cite{neftci2019}, perform slightly better with 89.93\% accuracy, and our model is very close to it. 

The K-MNIST results further reinforce the strength of the proposed model, which achieves an accuracy of 91.26\%, exceeding Ororbia’s~\cite{csdp2024} significantly denser FF-based SNN model that employs 7000 neurons in each of its two hidden layers of architecture and achieves 91.14\%. This comparison highlights the efficiency of the proposed architecture, which utilizes only two hidden layers with 500 neurons each. Moreover, the proposed model substantially outperforms the model by Terres et al.~\cite{snnffrobust} with the accuracy of 85.91\%, which also uses the FF approach but with a wider architecture of two hidden layers with 1400 conventional LIF neurons each.

In the N-MNIST dataset, our model achieves competitive results with 97.26\% accuracy using the LIF neuron model in the two hidden layers of 500 neurons architecture. 
Perez-Nieves et al.~\cite{sparse2021} use a single hidden layer of 200 neurons, achieving 92.70\%, and Taylor et al.~\cite{oxford2023} achieve 95.91\% accuracy, showcasing the adaptability of our model. Neftci et al.~\cite{neftci2019}, using LIF$^{\beta}$ neurons and a similar architecture, achieves the highest accuracy at 97.47\%. These results highlight strong performance for our model, although it slightly underperforms relative to Neftci et al.’s model.

In the SHD dataset, our model, with two hidden layers of 500 neurons and recurrent LIF architecture, achieves a solid 77.87\% accuracy. Cramer et al.~\cite{cramerSHD} and Neftci et al.~\cite{neftci2019} use recurrent LIF neurons with smaller architectures, each employing one hidden layer of 128 and 300 neurons, respectively, reaching 71.4\% and 70.81\%. The higher performance of our model demonstrates its strength in handling time-series data.

The table highlights that our model demonstrates strong performance across multiple datasets, particularly excelling on MNIST with state-of-the-art accuracy while also performing very competitively on FMNIST and N-MNIST, and on temporal datasets like SHD, our model achieves remarkable results using a recurrent LIF architecture and outperforms other BP-based models.
\begin{table*}[t]
\caption{Performance of our SNN model compared to the existing ANN implementation of the Forward-Forward algorithm}
\label{table:results2}
\begin{center}
\resizebox{0.85\textwidth}{!}{
\renewcommand{\arraystretch}{1.3}
\begin{tabular}{cccccccccc}
\hline
\multicolumn{1}{c}{\bf Dataset}  &\multicolumn{1}{c}{\bf Model} &\multicolumn{1}{c}{\bf Architecture} &\multicolumn{1}{c}{\bf Network Type} &\multicolumn{1}{c}{\bf Accuracy (\%)} \\
\hline

% MNIST
\multicolumn{1}{c}{\multirow{2}{*}{MNIST}}
	& \textbf{Proposed model} &784-2000-2000 &SNN &$98.69$ \\
        & Ororbia et al.~\cite{Ororbia2023}  &784-2000-2000 & ANN  &98.66 \\
        & Hinton~\cite{hinton2022}  &784-2000-2000-2000-2000 & ANN  &98.63 \\
	& Gandhi et al.~\cite{gandhi2023} &784-2000-2000-2000-2000 & ANN  &98.63 \\
        & Aminifar et al.~\cite{Aminifar2024}  &784-2000-2000-2000-2000 & ANN  &98.60 \\
        & Ghader et al.~\cite{ghader2024}  &784-500-500 & ANN  &98.58 \\
        & Lee et al.~\cite{symba2023}  &784-2000-2000-2000 & ANN  &98.58 \\
	& Brenig et al.~\cite{brenig2023} &784-2000-2000-2000-2000 & ANN &98.03 \\
	& Tang.~\cite{tangff2023}  &784-200-200-200-50-50 &ANN &$98.00$ \\
        & Tosato et al.~\cite{Tosato2023}  &784-784-784-784 & ANN  &94.00 \\
	
    \hline
   
% F-MNIST
\multicolumn{1}{c}{\multirow{2}{*}{F-MNIST}}
	%&\textbf{Proposed model}  &784-500-500 &SNN &89.13 \\
	&\textbf{Proposed model}  &784-2000-2000 &SNN &90.27 \\
        & Ghader et al.~\cite{ghader2024}  &784-500-500-500 & ANN  &90.22 \\
	& Ororbia et al.~\cite{Ororbia2023}  &784-2000-2000 & ANN  &89.60 \\
	& Brenig et al.~\cite{brenig2023}  &784-2000-2000-2000-2000 & ANN &87.31 \\
	& Tosato et al.~\cite{Tosato2023}  &784-784-784-784 & ANN  &82.60 \\

    \hline

% CIFAR10
\multicolumn{1}{c}{\multirow{2}{*}{CIFAR-10}}
&\multicolumn{1}{c}{\multirow{1}{*}{\textbf{Proposed model}}}  &3072-2000-2000 &SNN & 54.03 \\

    & Aminifar et al.~\cite{Aminifar2024}  &3072-2000-2000-2000-2000 & ANN  &53.95 \\
    & Ghader et al.~\cite{ghader2024}  &3072-500-500-500 & ANN  &53.31 \\
	& Brenig et al.~\cite{brenig2023}  &3072-2000-2000-2000-2000 & ANN & 47.60 \\

\hline

\end{tabular}}
\end{center}
\end{table*}

%%%%%%%%%%%%%%%%%%%%%%%%%%%%%%%%%%%%%%%%%%%%%%%%%%%%%
After comparing the performance of our model with other models based on spiking neural networks, we also conducted a comparison with some models based on the Forward-Forward algorithm that were based on artificial neural networks.
Table~\ref{table:results2} compares the performance of various FF-based models across three datasets—MNIST, F-MNIST, and CIFAR-10—using different architectures and network types. The datasets are organized into sections, and for each, the accuracy of our 2-layer SNN model is compared against other referenced works that use ANN models. The number of neurons in the layers of our architectures is selected based on commonly reported architectures for each dataset.
\begin{figure*}[t]
\centering
   \begin{subfigure}{0.45\linewidth}
   \centering
   \includegraphics[width=\linewidth]{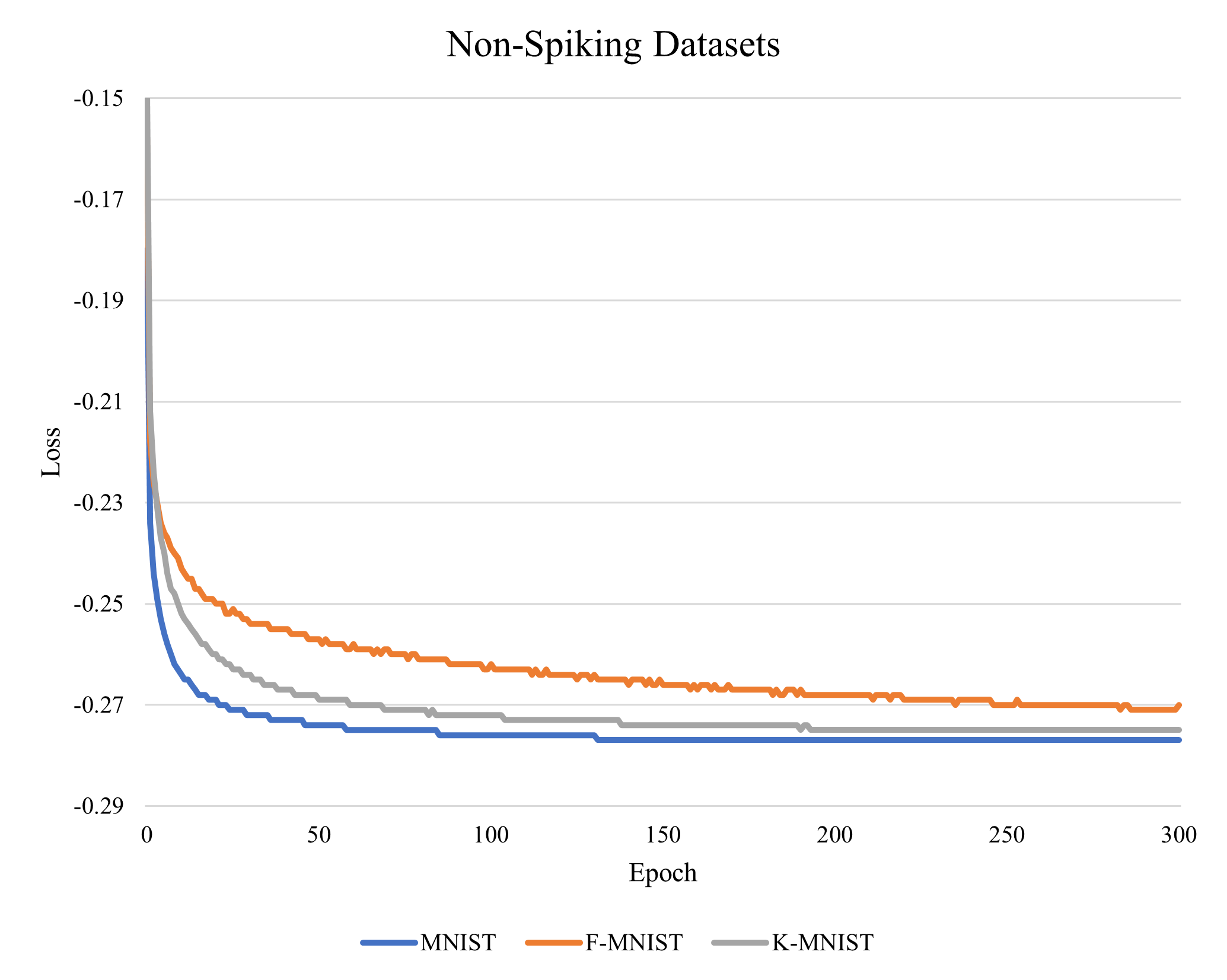}
   \caption{}
   \label{fig:loss1} 
\end{subfigure}
%\hfill
\begin{subfigure}{0.45\linewidth}
   \centering
   \includegraphics[width=\linewidth]{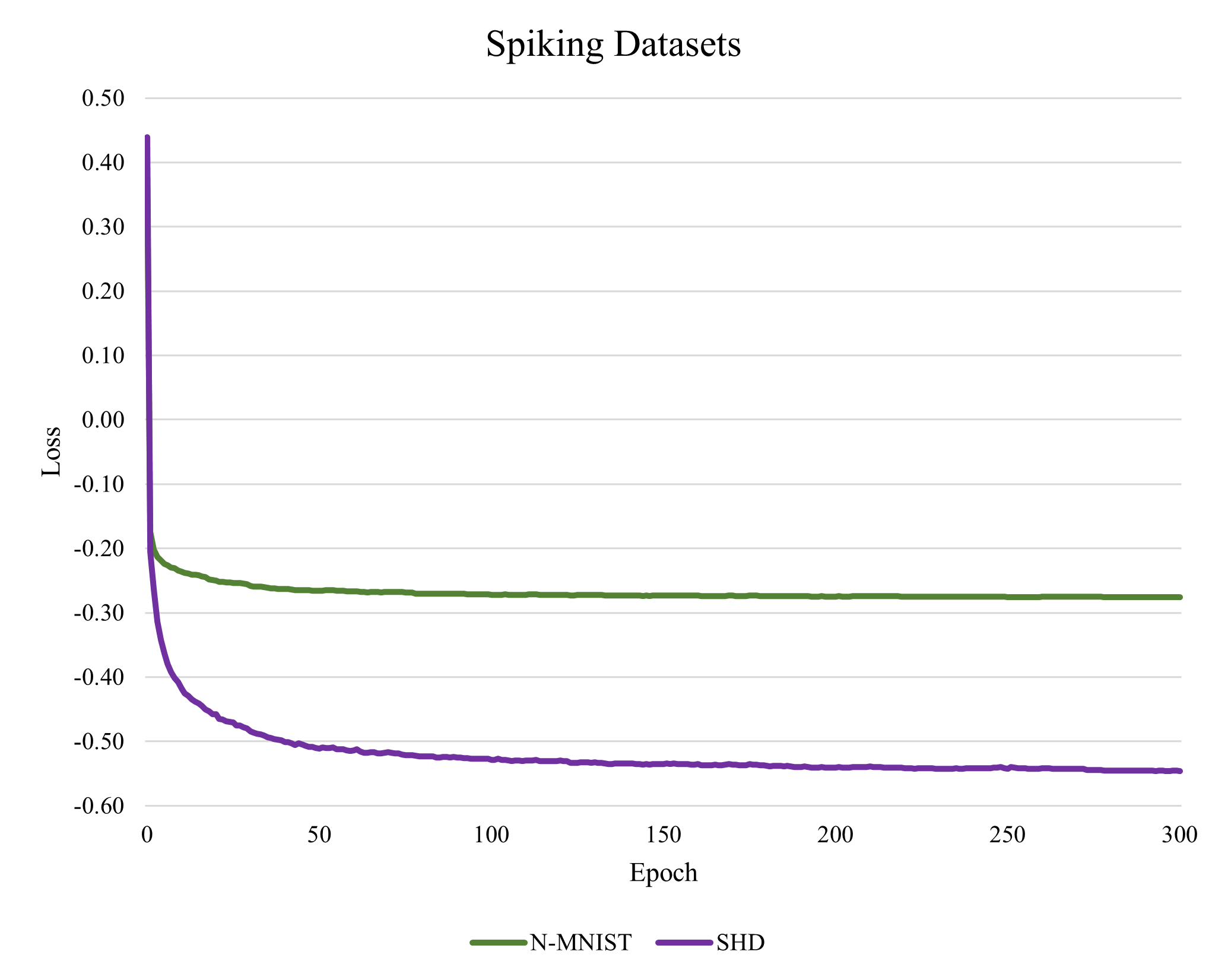}
   \caption{}
   \label{fig:loss2}
\end{subfigure}
\caption{Loss trend during training of our proposed FF-based spiking neural network model through epochs. (a) Loss changes for non-spiking datasets such as MNIST, F-MNIST, and K-MNIST. (b) Loss changes for spiking datasets such as N-MNIST and SHD.}
\label{fig:losses}
\end{figure*}
\begin{figure*}[t]
\centering
   \begin{subfigure}{\linewidth}
   \centering
   \includegraphics[width=\linewidth]{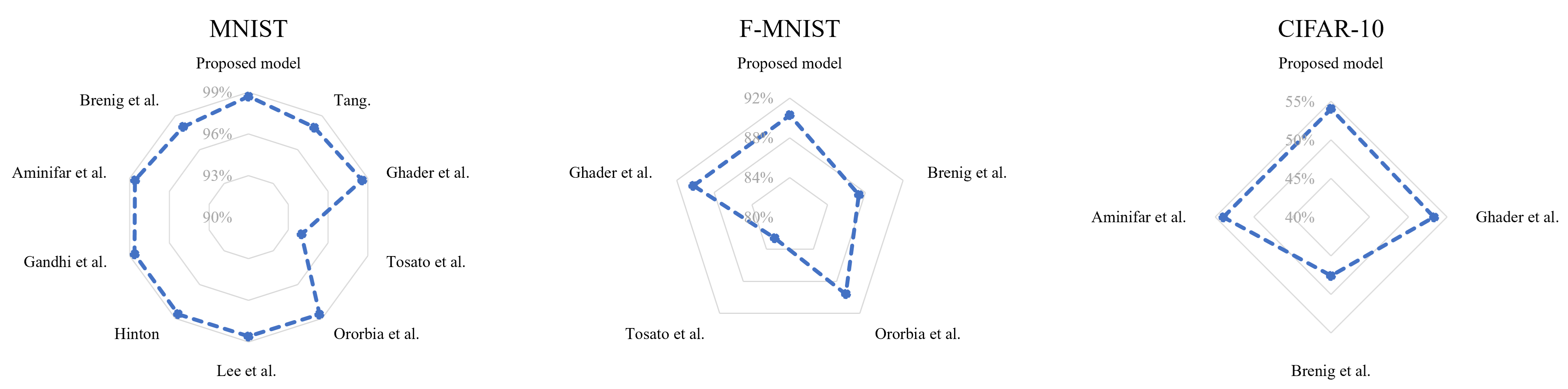}
   \caption{}
   \label{fig:dragratio} 
\end{subfigure}
\\[\baselineskip]
\begin{subfigure}[H]{\linewidth}
   \centering
   \includegraphics[width=\linewidth]{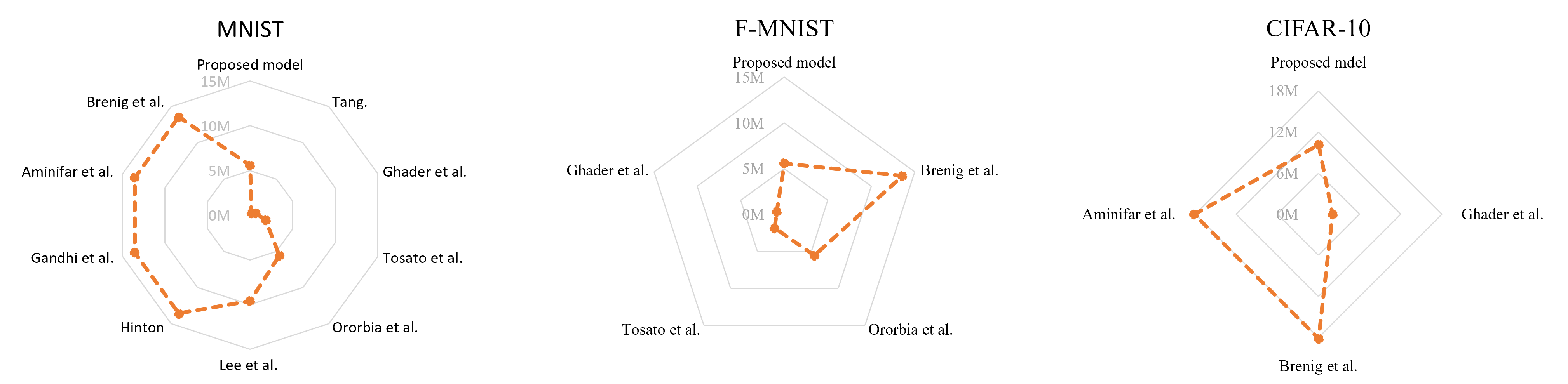}
   \caption{}
   \label{fig:dragratio3}
\end{subfigure}
\centering
\caption{Comparison of the accuracy and parameter count between our proposed SNN model and other ANN models based on the Forward-Forward algorithm. (a) Accuracy of our proposed model compared to other FF-based models trained with ANN. (b) Parameters count of our proposed model compared to other FF-based models trained with ANN.}
\label{fig:radar}
\end{figure*}
On the MNIST dataset, the proposed SNN model achieves a peak accuracy of 98.69\%, surpassing compared ANN-based models, including many with substantially deeper architectures. Remarkably, it marginally outperforms the models of Ororbia et al.~\cite{Ororbia2023}, Hinton~\cite{hinton2022}, Gandhi et al.~\cite{gandhi2023}, and Aminifar et al.~\cite{Aminifar2024}, each of which reports an accuracy of 98.66\%, 98.63\%, 98.63\%, and 98.60\%, respectively. This is achieved despite the proposed model employing a comparatively shallower architecture, consisting of two hidden layers with 2000 neurons each, as opposed to the four hidden layers used in the models by Hinton~\cite{hinton2022}, Gandhi et al.~\cite{gandhi2023}, and Aminifar et al.~\cite{Aminifar2024}. 

For the F-MNIST dataset, which presents a more challenging classification task than MNIST due to greater intra-class variability, the proposed SNN model once again achieves the highest accuracy among existing ANN-based models trained using the FF algorithm, reaching 90.27\%. This performance slightly surpasses that of the model proposed by Ghader et al.~\cite{ghader2024}, which achieves 90.22\%, and significantly outperforms other FF-trained ANN models, including those of Ororbia et al.~\cite{Ororbia2023}, Brenig et al.~\cite{brenig2023}, and Tosato et al.~\cite{Tosato2023}, which report accuracies of 89.60\%, 87.31\%, and 82.60\%, respectively. These findings underscore the robustness of the proposed model across varying data distributions and its capacity to generalize effectively in classification scenarios.

The CIFAR-10 dataset serves as a more complex benchmark, consisting of natural, non-spiking image data with high-dimensional color inputs. While the increased task difficulty leads to a reduction in overall accuracy, the proposed SNN model achieves a performance of 54.03\%, the highest among the compared ANN-based FF models. It slightly outperforms the deep ANN model introduced by Aminifar et al.~\cite{Aminifar2024}, which achieves 53.95\%, and significantly exceeds the results reported by Ghader et al.~\cite{ghader2024} and Brenig et al.~\cite{brenig2023}, which obtain 53.31\% and 47.60\%, respectively. Considering the non-spiking nature and high-dimensional input of CIFAR-10, the ability of the proposed SNN to match or surpass ANN-based counterparts represents a notable achievement and further demonstrates the potential of the forward-forward learning paradigm in complex visual recognition tasks.
These findings suggest that the proposed SNN architecture, when trained with the Forward-Forward learning rule, not only matches but often exceeds the performance of conventional ANN-based Forward-Forward models across diverse datasets. Its efficiency, driven by a relatively shallow design and biologically inspired neuron dynamics, points to the promise of SNNs as viable and adaptable alternatives to standard deep learning frameworks, particularly in contexts where backpropagation-free learning is desirable.

One of the points that is considered in training deep neural network models is analyzing the training process. Figure~\ref{fig:losses} shows the loss changes during our FF-based spiking neural network model training on both non-spiking and spiking datasets. The x-axis represents the epochs, while the y-axis represents the loss value.
For the non-spiking datasets, the loss for MNIST starts at a value and gradually decreases, stabilizing after around 100 epochs. The Fashion-MNIST dataset follows a similar pattern but reaches a slightly lower final loss than MNIST. The CIFAR-10 dataset shows more fluctuations, indicating that the model struggles more with this dataset compared to the others.
For the spiking datasets, the loss for N-MNIST starts at a moderately low value and decreases gradually, stabilizing around -0.3. The SHD dataset exhibits a much steeper initial loss decrease and stabilizes at a lower loss value compared to N-MNIST. By evaluating the overall changes in the loss values obtained in model training, it appears that the presented model has a good learning ability and has converged after performing certain epochs. There is this point that when faced with more complex datasets such as CIFAR-10, the process of large changes is not always smooth and decreasing, which is expected to be improved by using different architectures.

Considering the architectures and results obtained in Table~\ref{table:results2}, a relative comparison can be made between the number of parameters used in the models and their performance. Figure~\ref{fig:radar} compares our SNN model and other ANN models, both based on the Forward-Forward algorithm, across the MNIST, F-MNIST, and CIFAR-10 datasets. In the MNIST dataset, our proposed SNN model, with 5.57 million parameters, ranks third in terms of the number of network parameters and has the best performance compared to other models. The ANN model that we presented earlier in the study~\cite{ghader2024} had 0.64 million parameters, which could be among the selected models if the goal is to select the best model with a smaller parameter volume. A similar situation occurs in the F-MNIST dataset, where our SNN model achieves the highest accuracy while ranking third in parameter count. Therefore, if reducing the number of parameters is a priority and a slight trade-off in accuracy is acceptable, the model used in~\cite{ghader2024} could be a viable alternative.
On the CIFAR-10 dataset, our SNN model achieves higher accuracy than all compared ANN models. In terms of parameter efficiency, the ANN model we introduced earlier in this study~\cite{ghader2024} offers a favorable balance between performance and model size. However, our proposed SNN model, with approximately 10.14 million parameters, uses significantly fewer parameters than the other evaluated models, which average around 18.14 million. These comparisons highlight that our model not only delivers the best performance among the evaluated approaches but also maintains a competitive parameter count.

Another important consideration in spiking neural networks is the number of time steps required to train the network, as this parameter directly affects both computational cost and inference latency. Longer time steps typically allow the network to accumulate more information and stabilize spike-based learning, potentially leading to higher accuracy; however, this comes at the expense of increased energy consumption and prolonged training time. Conversely, reducing the number of time steps can enhance efficiency and inference speed, though it may compromise learning performance if the network lacks sufficient temporal resolution to capture salient features.
Inference in SNNs involves multiple feedforward computations, each corresponding to a time step, and relies on discrete and sparse spike activity rather than continuous activations. When backpropagation is used for training, spike signals often weaken over time, resembling the vanishing gradient problem observed in traditional ANNs. To prevent the loss of important information, SNNs frequently require a relatively large number of time steps, often in the triple digits~\cite{sstdp}.
In contrast, our model employs only 10 time steps during training across all evaluated datasets. This relatively small number of time steps, compared to the majority of other SNN models, highlights the efficiency of our proposed algorithm and its ability to achieve competitive performance while maintaining a significantly lower computational burden in terms of the number of time steps.

%--------------------------------------Conclusion-----------------------
\section{Conclusion}
\label{sec5}
This paper presented a novel approach to training spiking neural networks using the Forward-Forward algorithm. Unlike traditional backpropagation, the FF algorithm offers a more biologically plausible training method relative to BP by utilizing two forward passes instead of the typical forward and backward pass. This approach addresses key limitations of backpropagation, such as the need for global feedback pathways, explicit storage of all intermediate activities, and weight symmetry, all of which are biologically implausible.
Our experimental results demonstrated the effectiveness of the FF algorithm in training SNNs across some spiking (Neuro-MNIST, SHD) and non-spiking (MNIST, Fashion-MNIST, K-MNIST) datasets. The proposed FF-based SNN model achieved competitive accuracy, often outperforming traditional backpropagation-based SNNs.
The compatibility of the Forward-Forward algorithm with spiking-based architectures enhances its suitability for deployment in practical settings, especially those that impose constraints on power efficiency and require rapid, online learning capabilities. Moreover, the algorithm’s inherent advantages—such as layer-wise local learning and its ability to function effectively with black-box components—position it as a promising framework for future advancements in specialized domains, particularly neuromorphic computing. Future research could focus on optimizing the FF algorithm, integrating it with emerging neuromorphic platforms, and applying it to increasingly complex and diverse tasks, thereby enabling the development of more robust and widely applicable spiking neural network solutions.

\end{document}